\documentclass{article}
\usepackage[dvipdfmx]{graphicx}
\usepackage{epstopdf}
\usepackage{color}
\usepackage[a4paper, left=0.5in, right=0.5in, top=0.75in, bottom=0.75in]{geometry}
\usepackage{amsmath,amssymb}
\usepackage{makecell}
\usepackage{textcomp,marvosym}
\usepackage{nameref}
\usepackage[hidelinks]{hyperref}
\usepackage{array}
\usepackage{adjustbox}
\usepackage{makecell}
\usepackage{amssymb}
\usepackage{pifont}
\usepackage{multicol}
\usepackage[right]{lineno}
\usepackage[nopatch=eqnum]{microtype}
\DisableLigatures[f]{encoding = *, family = * }
\usepackage[table]{xcolor}
\usepackage{array}
\newcolumntype{+}{!{\vrule width 2pt}}
\newlength\savedwidth

\raggedright
\usepackage[aboveskip=1pt,labelfont=bf,labelsep=period,justification=raggedright,singlelinecheck=off]{caption}

\makeatletter
\renewcommand{\@biblabel}[1]{\quad#1.}
\makeatother

\usepackage{cleveref}
\usepackage{enumitem}

\crefname{mytype}{Test Box}{Test Boxes}
\usepackage[]{mdframed}
\usepackage[most]{tcolorbox}
\newtcolorbox[use counter=mytype]{testbox}[2]{
  enhanced,
  label type=mytype,
  label = #1,
  title = Test~#1:~#2
}

\newtcolorbox{myboxVS}[2][]{
  enhanced,
  attach boxed title to top left={yshift=-0.3\baselineskip-0.4pt,xshift=2mm},
  colframe=black,
  colback=white,
  fonttitle=\bfseries,
  colbacktitle=white,
  coltitle=black,
  boxed title style={
    boxrule=0pt,
    colframe=white,
    },
  title=#2,
  #1, breakable}

\newtcolorbox{myframe}[2][]{%
  enhanced,colback=white,colframe=black,coltitle=black,
  sharp corners,boxrule=0.4pt,
  fonttitle=\itshape,
  attach boxed title to top left={yshift=-0.3\baselineskip-0.2pt,xshift=3mm},
  boxed title style={tile,size=minimal,left=0.5mm,right=0.5mm,
    colback=white,before upper=\strut},
  title=#2,#1
}

\usepackage{wrapfig}
\usepackage{pdflscape}
\usepackage{longtable}
\usepackage{pdfpages}
\usepackage{float}
\usepackage{xcolor}
\usepackage[normalem]{ulem}
\usepackage{lmodern}
\usepackage[T1]{fontenc}

\lstdefinestyle{jsonstyle}{
    backgroundcolor=\color{white},   
    basicstyle=\ttfamily\small,      
    breaklines=true,                  
    showstringspaces=false,           
    commentstyle=\color{green},       
    keywordstyle=\color{blue},        
    stringstyle=\color{red},          
    numbers=left,                     
    numberstyle=\tiny\color{gray},    
    stepnumber=1,                     
    numbersep=5pt                     
}
\crefname{listing}{Program-code}{Program-codes}  
\Crefname{listing}{Program-code}{Program-codes}

\usepackage[numbers]{natbib}
\title{\bf Language Model Powered Digital Biology with BRAD}

\author{Joshua Pickard$^{1}$,
Ram Prakash$^{1}$,
Marc Andrew Choi$^{2}$,
Natalie Oliven$^{1}$, Cooper Stansbury$^{1}$,\\
Jillian Cwycyshyn$^{3}$,
Alex Gorodetsky$^{4}$, \\ 
Alvaro Velasquez$^{5}$,
and Indika Rajapakse$^{1,2,6}$\\
}

\date{}

\begin{document}

\maketitle

\noindent
{\small $^{1}$Department of Computational Medicine and Bioinformatics, University of Michigan, Ann Arbor, MI, USA
\small $^{2}$Department of Mathematics, University of Michigan, Ann Arbor, MI, USA
\small $^{3}$Department of Biomedical Engineering, University of Michigan, Ann Arbor, MI, USA
\small $^{4}$Department of Aerospace Engineering, University of Michigan, Ann Arbor, MI, USA
\small $^{5}$Department of Computer Science, University of Colorado Boulder, Boulder, CO, USA
\small $^{6}$Corresponding authors: \texttt{\{indikar\}@umich.edu}}

%
%

\vspace{3mm}
\noindent
\textbf{Motivation:}
Recent advancements in Large Language Models (LLMs) are transforming biology, computer science, engineering, and every day life.
However, integrating the wide array of computational tools, databases, and scientific literature continues to pose a challenge to biological research.
LLMs are well-suited for unstructured integration, efficient information retrieval, and automating standard workflows and actions from these diverse resources.
To harness these capabilities in bioinformatics, we present a prototype Bioinformatics Retrieval Augmented Digital assistant (BRAD).

\vspace{1mm}
\noindent
\textbf{Results:}
BRAD is a chatbot and agentic system that integrates a variety of bioinformatics tools.
The Python package implements an AI \texttt{Agent} that is powered by LLMs and connects to a local file system, online databases, and a user's software.
The \texttt{Agent} is highly configurable, enabling tasks such as Retrieval-Augmented Generation, searches across bioinformatics databases, and the execution of software pipelines.
BRAD's coordinated integration of bioinformatics tools delivers a context-aware and semi-autonomous system that extends beyond the capabilities of conventional LLM-based chatbots.
A graphical user interface (GUI) provides an intuitive interface to the system.
\vspace{1mm}

\noindent
\textbf{Availability and Implementation:} The source code for BRAD is available at \href{https://github.com/Jpickard1/BRAD}{\textcolor{blue}{\texttt{https://github.com/Jpickard1/BRAD}}}, and the GUI distribution can be accessed via \href{https://hub.docker.com/r/thedoodler/brad?uuid}{\textcolor{blue}{\texttt{docker hub}}}.
Additional information for installation, tutorials, documentation, and further development can be found on: \href{https://brad-bioinformatics-retrieval-augmented-data.readthedocs.io/en/latest/}{\textcolor{blue}{\texttt{ReadTheDocs}}} and in the \href{https://brad-bioinformatics-retrieval-augmented-data.readthedocs.io/_/downloads/en/latest/pdf/}{\textcolor{blue}{\texttt{software manual}}}.

\begin{multicols}{2}
\section{Introduction}
Recent advancements in artificial intelligence (AI) are revolutionizing the use of large language models (LLMs) in biomedical research and bioinformatics.
From automating laboratory experiments \cite{boiko2023autonomous} to virtual laboratories \cite{swanson2024virtual} and analyzing single-cell data \cite{schaefer2024multimodal}, LLMs are transforming how researchers approach complex tasks.
While much of LLM development has been driven by industry, the research community has unique needs, such as up-to-date literature access and tailored data pipelines.
For example, retrieval-augmented generation (RAG) enables LLMs to access the most recent literature and generate verifiable answers \cite{lewis2020retrieval, craig2024lmrac, matsumoto2024kragen}.
Equipping LLMs with research tools and databases provides autonomy and transparency, capabilities that general-purpose models cannot replicate.
However, deploying LLM applications in research ecosystems requires overcoming additional challenges associated with handling computational infrastructure, interface design, and integration with existing tools.

We developed a Bioinformatics Retrieval Augmented Digital assistant (BRAD) to streamline the deployment of LLMs with biomedical research tools. 
BRAD provides a chat interface to a LLM powered \texttt{Agent} capable of information retrieval from user-provided documents, online literature repositories, software pipelines, and public databases.
The agent-based design enables integration with a wide range of computational tools and resources, extending its capabilities beyond those of other RAG systems \cite{craig2024lmrac, matsumoto2024kragen}.
The modular architecture includes a flexible framework for integrating data from additional sources.

\section{Software Architecture}
\begin{figure*}[t!]
    \centering
    \includegraphics[width=0.9\linewidth]{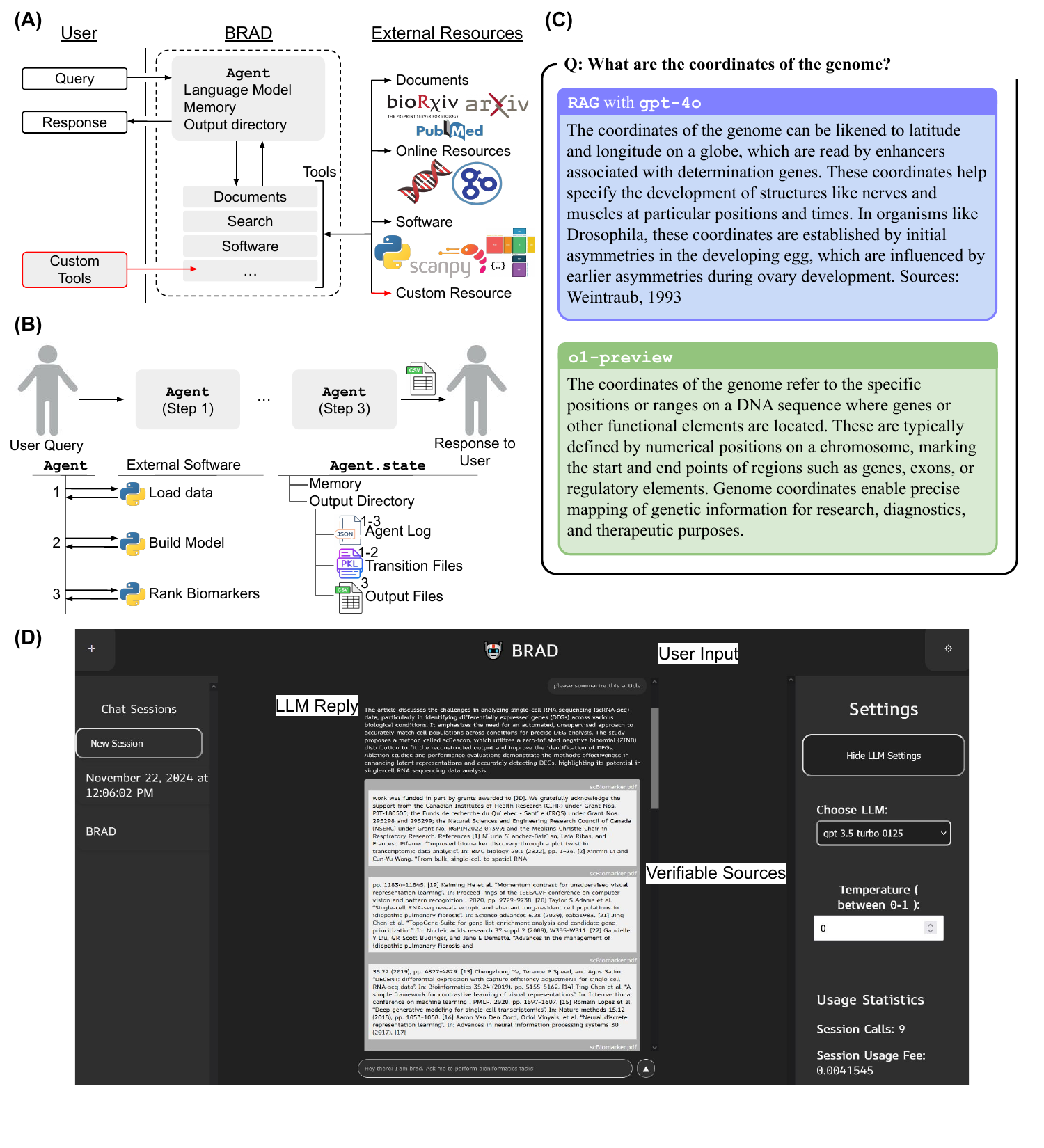}
    \vspace{-10mm}
    \caption{\textbf{Overview of BRAD’s architecture, capabilities, and user interface.} (A) The modular architecture uses the \texttt{Agent} as an interface between the user, the LLM, and tools that access external resources. The document and online databases are deployed in the GUI, and the modular design allows users to add new tools, similar to the software tool, to interface and retrieve information from new sources. (B) Agentic workflows utilize multiple tools to address user queries. In the biomarker workflow, the \texttt{Agent} uses the Software tool to run Python scripts, each producing an output file written to the output directory and updates the \texttt{Agent}'s memory. The output from one stage serves as input for the next, culminating in an interpretable spreadsheet for the user. Additional software can be integrated to support tasks beyond biomarker selection. (C) Example outputs with the newest commercial LLM versus a RAG pipeline. (D) The Graphical User Interface (GUI) of BRAD highlights the user query, the LLM response, and the retrieved information. The left panel enables users to switch between chat sessions, while the right panel provides options to adjust settings for the LLM, the RAG pipeline, and additional features.}
    \label{fig: main figure}
\end{figure*}

The BRAD Python package provides a flexible platform for automating bioinformatics workflow through the integration of LLMs (\cref{fig: main figure}A).
An \texttt{Agent} class -- a virtual AI system with memory and interaction capabilities -- orchestrates interactions between core and tool modules \cite{langgraph, yao2022react, wu2023autogen}.
Core modules manage the \texttt{Agent}'s memory and resource utilization, such as LLM operations, while tool modules connect the \texttt{Agent} to external resources like databases, literature repositories, and software.
To enhance usability, the package can be deployed within Python environments, as a command-line interface, or through a Graphical User Interface (GUI).
First, we overview of the Python package's architecture and then explain the tool modules and GUI deployment.

\subsection{Python Architecture}
The \texttt{Agent} class serves as the user interface with BRAD.
It coordinates the \texttt{Agent}'s memory, manages the output directory, and oversees the utilization of system resources, including LLMs and external tools.
The interfaces to these tools are standardized, enabling an \texttt{Agent} to be configured with different or new tools.

Tool modules extend the \texttt{Agent}'s functionality by providing an interface to connect external databases, software, or other resources.
These modules are designed following the RAG paradigm, where the LLM generated reply is augmented by retrieval of relevant information from external data sources.
The modular design allows users to configure their \texttt{Agent} with tools tailored to specific needs.
Moreover, the tool interface is minimal, ensuring that new resources can be easily integrated into the framework.

BRAD is configured to integrate a broad range of LLMs.
Users can leverage locally hosted models, which can be fine-tuned for domain-specific tasks, or connect to online models such as \href{https://build.nvidia.com/meta/llama3-70b}{NVIDIA's NIM platform} and \href{https://openai.com/}{OpenAI}'s offerings.
BRAD is built using LangChain ensuring compatibility and to facilitate the ability to integrate with other standard frameworks \citep{affarth2023generative}.

\subsection{Tool Modules}

\subsubsection{Document Chat Tool}
The Document Chat module retrieves information from documents, including research articles or any PDF, using a Retrieval Augmented Generation (RAG) pipeline \citep{lewis2020retrieval, shuster2021retrieval}.
RAG enhances the generation process by supplying the LLM with verifiable, external information that was not used to train the LLM.

This module contains several methods to perform retrieval with similarity, multi-query, and maximal marginal relevance retrievers.
It enhances the retrieved texts by applying contextual compression, including surrounding context of the retrieved text, and incorporating source information, among other techniques. 
The retrieved information is incorporated into a prompt template, such as the following, which is then passed to the LLM \cite{sahoo2024systematic}:
\begin{quote}
\noindent
\texttt{\textbf{System:} Respond to the human based on the provided sources.} \\
\texttt{\textbf{Sources:} \{Retrieved and augmented text is placed here.\}} \\
\texttt{\textbf{Human:} \{User input is placed here.\}}
\end{quote}
Providing explicit, well-sourced information to the LLM improves the quality of generated responses by: (1) minimizing hallucinations, (2) anchoring replies in reliable data, and (3) allowing users to verify the response against the cited sources.

As an example, consider the open ended question (\cref{fig: main figure}C): What are the coordinates of the genome?
We answered this using OpenAI's \texttt{o1-preview} without RAG and \texttt{gpt-4o} with RAG.
The \texttt{o1} model responds like a contemporary bioinformatician, focusing on genomic loci and their numerical representations, as commonly used when aligning assays such as Hi-C.
In contrast, the RAG-enabled system offers a more metaphorical response, likening the nucleus to a world or globe — a perspective that once dominated biological thought \cite{weintraub1993summary}.
Both answers are scientifically valid.
While answer preferences vary, the RAG-generated response has the added benefit of referencing a specific source, grounding its perspective in the context of prior work.

\subsubsection{Search Tool}
The Search tool, like RAG, retrieves information from online databases that are external to BRAD.
As of this writing, the following sources are searchable with BRAD: arXiv, PubMed, and bioRxiv.
Users are given options to download retrieved search results and include them in a RAG database.
Similarly, BRAD's search tool was extended to search bioinformatics specific databases including those found in Enrichr and Gene Ontology.
The system can take user queries and identify search terms, or it can upload genes to perform enrichment on from data files.

Custom prompt templates facilitate literature and database searches by leveraging the LLM to select the appropriate database and search terms.
An example of the prompt template is as follows:
\begin{quote}
\noindent
\texttt{\textbf{System:} Select an appropriate database for the user's query.} \\
\texttt{\textbf{Database Descriptions:} \{ARXIV, BIORXIV, PUBMED, ...\}} \\
\texttt{\textbf{Human:} \{User input is placed here.\}}
\texttt{\textbf{Output}:}\\
1. \texttt{<Enter a database name>}\\
2. \texttt{<Enter search terms>}
\end{quote}
After searching the selected online databases, the LLM generates a response to the user using a prompt template, like the RAG pipeline.
Both the Document Chat and Search tools are integrated with both the Python package and the GUI.

\subsubsection{Module Template}
To ensure modularity, BRAD has a standard template for tool modules.
Each module exclusively interacts with \texttt{Agent.state}, which provides access to the LLM, user data, system memory, and more.
Standardized output and logging conventions ensure uniformity and proper updates to \texttt{Agent.state} during tool usage.
Supporting any LangChain compatible LLM, BRAD tool modules can leverage the full range of retrievers and tools available in the LangChain and LangGraph frameworks \cite{affarth2023generative}.

\subsection{Deployment and Distributions}

\subsubsection{Graphical User Interface (GUI)}
The BRAD GUI offers a chatbot interface for interacting with the \texttt{Agent} class.
A Node.js frontend communicated with a Flask API server that is built atop of the Python package.
The Flask API, contains endpoints to \texttt{/invoke} the \texttt{Agent} class and query the BRAD chatbot; create, delete, and manage chatbot \texttt{/sessions} and RAG \texttt{/databases}; select the usage of different \texttt{/llm}s; and manipulate different settings or \texttt{/configurations}.

In BRAD's GUI, the user can upload their own documents to construct new RAG databases and also manipulate the retrieval and augmentation methods.
When BRAD uses RAG to answer a question, the user can click the \texttt{Agent}'s reply to reveal which documents were used to generate the reply and automatically open the associated documents (\cref{fig: main figure}D).

\subsubsection{Deployment Options}
To bridge the gap between the development and users in our laboratory, we implemented a continuous integration and continuous development (CI/CD) pipeline.
This pipeline enables near real-time distribution of software updates and new tools, ensuring efficient deployment.

\section{Results and Applications}

\subsection{Biomarker Identification Workflow.}
Identifying novel biomarkers is a common research task with broad applications \cite{califf2018biomarker}.
We developed a Software tool that interfaces an \texttt{Agent} with software external to BRAD, including a biomarker selection pipeline.
This module follows a workflow similar to RAG, where the LLM is prompted with information retrieved from software documentation and generates the code required to run the software.
While generating executable code with LLMs remains an open and active area of research, enabling LLM agents to execute predefined code is a more constrained and achievable objective.

We deployed an \texttt{Agent} with the Software tool and provided access to the multistage biomarker identification workflow of \cite{pickard2024biomarker}.
The \texttt{Agent} manages a queue of actions and tools to execute the multistage biomarker workflow for a single user query, using the output directory and internal memory to transfer information and data between stages.
Instead of generating a sentence like a typical LLM, the \texttt{Agent} produces a spreadsheet from the software in the following format:
\begin{quote}
\noindent
\texttt{| Biomarker | $\ \ \ \ \lambda$ | Rank |}\\
\texttt{---------------------------------------------------------}\\
\texttt{| CDT1$\qquad\ \ $|$\ \ \ $ 10    |$\ \ \ $ 1    |}\\
\texttt{| PCNA$\qquad\ \ $|$\ \ \ \ $ 9    |$\ \ \ $ 2    |}\\
\end{quote}
The table is generated reproducibly using predefined methods specified by the external software.
This integration uniquely combines the capabilities of an LLM with a software pipeline.
An LLM alone provides only generic responses disconnected from user-specific data (\S\ref{SI subsec bmk o1preview comparison}), while the absence of an LLM requires users to possess significant expertise to execute the workflow.

Beyond the biomarker pipeline, the Software tool can interface an \texttt{Agent} to other external software.
We deployed this system to interface with the \texttt{AnnData} and \texttt{Scanpy} libraries for single cell data.
However, the use of this tool (1) requires the software executed by the \texttt{Agent} to have detailed documentation and (2) is susceptible to errors in LLM generated code.
While the system works robustly for the biomarker pipeline, because generating bug free code with LLMs remains an active area of research, this tool available in the Python package and not deployed in the GUI.

\subsection{Cost and Benchmarking BRAD's Tools}
We evaluated the costs associated with the Document Chat, Online Search, and Software tools, considering metrics such as runtime, API fees, and the number of LLM input/output tokens.
Notably, the integration of RAG does not significantly increase these costs compared to using the LLM alone.
Open-ended prompts to the LLM—such as those from the Document Chat or Software tools, which involve generating code or summarizing documents—incur higher costs than using the LLM to query database tools selectively.
For additional details, refer to \S\ref{SI section benchmarking tool modules}.

To assess the quality of output generated by the RAG pipeline, we employed the RAG Assessment framework \cite{es2023ragas}.
Using a database of 500 unique documents spanning 18,000 pages, the RAG pipeline — tested with both standard and advanced configurations — demonstrated notable improvements in answer relevance and faithfulness.
These results align with reduced hallucination rates observed in other RAG pipelines.
Further details are provided in \S\ref{SI Quantitative Results}.

\section{Summary}
BRAD is a prototype AI research assistant that integrates large language models (LLMs) with information retrieval from bioinformatics resources.
Its modular architecture and diverse data sources enhance the accuracy and transparency of LLM-generated responses while offering flexibility to incorporate additional tool modules.
Distributed as a Python package with a user-friendly GUI, BRAD is designed to be both extensible and easy to deploy, making it an adaptable research tool. 
BRAD is designed for both extensibility and ease of deployment, making it an adaptable and user-friendly research tool.
This positions BRAD as a valuable resource for researchers seeking to integrate LLMs into their workflows and broader research ecosystems.
\end{multicols}

%
%
\section*{Acknowledgments}
We thank Adam Lord of NVIDIA, Santosh Srivastava of General Electric, and members of the Rajapakse Lab for helpful and inspiring discussions.
We also thank Vivan Nyati, a student of Greenhills, Ann Arbor.
This work was supported by the Defense Advanced Research Projects Agency award number HR00112490472 (IR), the Air Force Office of Scientific Research (AFOSR) award number FA9550-22-1-0215 (IR), support from NVIDIA (IR), and  NIGMS GM150581 (JP).
BRAD and other LLMs proposed edits for clarity and grammar, with all final decisions made by the authors.

\bibliography{refs}

%
%
\newpage
\begin{center}
    \Large\bf Supporting Information
\end{center}

\section{Results and Examples}
This section presents extended results and details related to the use cases of BRAD discussed in the paper.

\subsection{Benchmarking Tool Modules}\label{SI section benchmarking tool modules}
\begin{figure}[h]
    \centering
    \includegraphics[width=\linewidth]{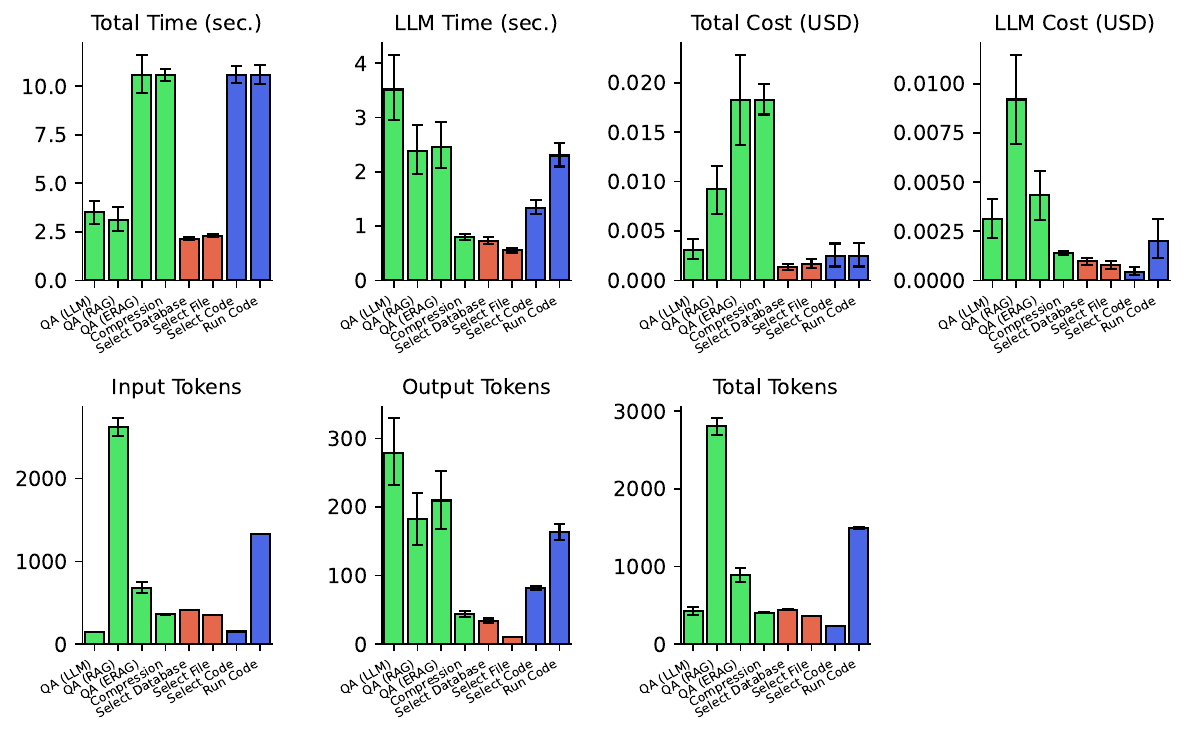}
    \caption{Costs associated with the use of LLM in each module of BRAD are profiled. Each task is color-coded according to its respective module: \texttt{LAB NOTEBOOK} is represented in green, \texttt{SOFTWARE} in red, and \texttt{DIGITAL LIBRARY} in blue.}
    \label{fig: benchmarking brad}
\end{figure}

This section profiles the use of the LLM in different modules of BRAD. The major cost to the user occurs in terms of (1) runtime and (2) API fees.
There is no substantial memory requirements for this system.
A user could also elect to run the LLM inference locally, in which case there would be no fees and the time requirements depend on the user's hardware. The following metrics are evaluated:
\begin{itemize}
    \item Total Time: the time required to respond to the users query by a single module.
    \item LLM Time: the time required for the LLM to perform its task within the module.
    \item Total Cost: the total cost of LLM API fees within a single module. This can represent multiple LLM utilizations.
    \item LLM Cost: the cost of LLM API fees for the LLM to perform its task within a module.
    \item Input Tokens: the number of tokens input for a single LLM query.
    \item Output Tokens: the number of tokens output for a single LLM query.
    \item Total Tokens: the sum of input and output tokens for a single LLM query.
\end{itemize}
We measured these seven metrics for user queries requiring BRAD to perform different tasks: engaging in question-answering (i.e., functioning purely as a chatbot), searching online databases (which involves selecting a database and potentially loading data from a file), and interacting with software by reading and writing software code and documentation.
See \cref{fig: benchmarking brad} and  \cref{tab: benchmarking figure pairs plot}.

\begin{table}[h]
\centering
\caption{Summary of cost of different tasks performed by LLMs.}\label{tab: benchmarking figure pairs plot}
\begin{tabular}{l|rrrr}
Task & Model & Total Tokens & LLM Time (sec.)& Total Cost (USD, $\times 10^{-4}$) \\
\hline\hline
QA (LLM)& \texttt{gpt-3.5-turbo-0125} & 303.14 ± 57.22 & 1.95 ± 0.67 & 0.0 ± 0.0 \\
 & \texttt{gpt-4o} & 680.18 ± 146.8 & 6.65 ± 1.71 & 0.009 ± 0.002 \\
\hline
QA (RAG) & \texttt{gpt-3.5-turbo-0125} & 2746.5 ± 340.04 & 1.67 ± 0.87 & 0.001 ± 0.0 \\
 & \texttt{gpt-4o} & 2863.77 ± 367.62 & 3.11 ± 1.58 & 0.017 ± 0.003 \\
\hline
QA (ERAG) & \texttt{gpt-3.5-turbo-0125} & 743.64 ± 202.12 & 1.5 ± 0.57 & 0.003 ± 0.0 \\
& \texttt{gpt-4o} & 1033.73 ± 372.96 & 3.41 ± 1.35 & 0.034 ± 0.007 \\
\hline
Compression & \texttt{gpt-3.5-turbo-0125} & 403.45 ± 73.0 & 0.73 ± 0.49 & 0.003 ± 0.0 \\
 & \texttt{gpt-4o} & 410.78 ± 77.89 & 0.87 ± 0.56 & 0.034 ± 0.007 \\
\hline
Call code & \texttt{gpt-3.5-turbo-0125} & 1488.21 ± 29.26 & 2.17 ± 0.52 & 0.001 ± 0.0 \\
 & \texttt{gpt-4o} & 1530.75 ± 51.25 & 3.22 ± 0.74 & 0.012 ± 0.001 \\
\hline
Select code & \texttt{gpt-3.5-turbo-0125} & 233.86 ± 7.0 & 1.25 ± 0.16 & 0.001 ± 0.0 \\
 & \texttt{gpt-4o} & 243.5 ± 14.15 & 1.88 ± 0.71 & 0.012 ± 0.001 \\
\hline
Select database & \texttt{gpt-3.5-turbo-0125} & 448.38 ± 29.29 & 0.78 ± 0.4 & 0.0 ± 0.0 \\
 & \texttt{gpt-4o} & 436.19 ± 34.02 & 0.59 ± 0.23 & 0.004 ± 0.001 \\
\hline
Select file & \texttt{gpt-3.5-turbo-0125} & 363.35 ± 1.69 & 0.47 ± 0.13 & 0.0 ± 0.0 \\
 & \texttt{gpt-4o} & 361.25 ± 1.65 & 0.71 ± 0.05 & 0.004 ± 0.0 \\
 \hline\hline
\textbf{Average} & \texttt{gpt-3.5-turbo-0125} & 575.45 & 1.02 & 0.002 \\
 & \texttt{gpt-4o} & 631.17 & 1.53 & 0.026 \\
\end{tabular}
\end{table}

\subsection{RAG Assessment (RAGAs) evaluation metrics}\label{SI Quantitative Results}
Five metrics were used to evaluate the utility of BRAD's RAG pipeline \cite{es2023ragas}:
\begin{itemize}
    \item Faithfulness: This quantifies the consistency between BRAD's answer and the literature.
    \item Answer Relevance: This evaluates if the response is applicable to the particular question.
\end{itemize}

Four components are used to calculate each of these metrics: question, response, context, and ground truth.
For an individual question, the context used to construct the question and the answer produced by BRAD's RAG are split into small, verifiable claims.
Then, statements from BRAD's answer can be compared in a one for one fashion to find claims indicated by BRAD substantiated or not substantiated in context of the literature.
This approach allows us to construct a standard confusion matrix where:
\begin{itemize}
    \item \textbf{TP:} True Positives are statements that are present in both the ground truth and the generated response.
    \item \textbf{FP:} False Positives are statements that are present in the generated response but not the ground truth.
    \item \textbf{TN:} True Negatives are statements that are present in the ground truth but not in the generated response.
    \item \textbf{FN:} False Negatives are statements that are present in the ground truth but not in the generated response.
\end{itemize}

\paragraph{Faithfulness.}
Faithfulness refers to the factual consistency of the generated response to a question against the given context.
Since this metric is entirely based on the context given, if the context given is faulty, then a factually incorrect answer can still be considered faithful.
Each of the generated claims is then checked against the given context to determine if the claim can be inferred from context.
\begin{equation*}
    \text{Faithfulness} = \frac{\text{Number of Claims from Response inferred from Context
}}{\text{Number of Claims from Response}}.
\end{equation*}

\paragraph{Answer Relevance.}
Answer Relevance is a measure of how closely the generated response and given prompt are related.
Answers that are incomplete or contain redundant information are typically assigned lower scores. We use the following formula to calculate answer relevance
\begin{equation*}
    \text{Answer Relevance} = \frac{1}{N} \left(\sum_{k=1}^N \frac{e_{g_k} \cdot e_{o}}{\|e_{g_k}\| \|e_{o}\|
}\right),
\end{equation*}
where $N$ is the number of generated questions, $e_{g_k}$ is the vectorized embedding of the $k$th generated question from a generated response, and $e_{o}$ is the vectorized embedding of the original question.

\begin{figure}[h]
    \centering
    \includegraphics[width=0.5\textwidth]{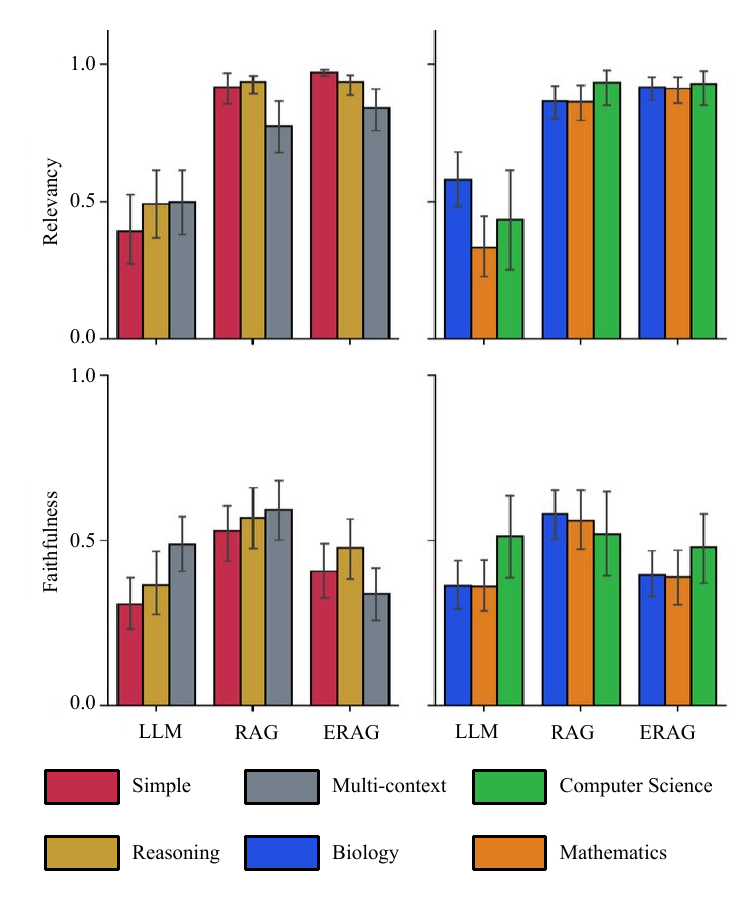}
    \caption{Visualizations of the RAGAs metrics using BRAD. The base LLM in all three BRAD bots is OpenAI GPT 4.o-mini, and the enhancements used in ERAG are Multiquery, Contextual Compression and Reranking.}
    \label{fig: RAGAs}
\end{figure}

\paragraph{Experiment.}
The performance of BRAD's RAG system is presented in \cref{fig: RAGAs}.
Notably, BRAD achieves higher performance metrics when utilizing the RAG pipeline compared to relying solely on the LLM.

\subsection{Biomarker Selection Extended Outputs and Analysis}\label{SI subsec bmk o1preview comparison}
To compare the utility of BRAD, using \texttt{gpt-3.5} as the LLM, versus the recently introduced \texttt{o1-preview} chatbot from OpenAI, we gave both systems the prompt:
\begin{quote}
    ``Execute a biomarker selection pipeline for RNAseq data."
\end{quote}
BRAD's Software tool directly interfaces with the data and biomarker software. This allows the system output (page 10) to summarize and synthesize results from the pipeline, including providing the top ranked genes directly to the user.
The biomarker software, a collection of python scripts that are external to BRAD and used by the \texttt{Agent}'s Software tool, can be accessed from the projects Github.

In contrast, \texttt{o1-preview}, a newer LLM than the one employed by BRAD, produced a seven-step framework for executing a biomarker selection workflow (page 11).
However, rather than addressing the specific user query, \texttt{o1-preview} offered general instructions on how such a task might be approached.
This response, while structured, is significantly less useful to users who require actionable results rather than abstract guidance.

A main challenge using LLMs for biomarker identification and other bioinformatics workflows is structuring data for input to a LLM.
Whereas BRAD used software documentation to run codes and interface with data from the Software tool, the \texttt{o1-preview} chatbot lacks agentic capabilities that allow users to interface the LLM with their system.
\texttt{o1-preview} can only accept text input.
To provide additional biological context to \texttt{o1-preview}, we used the abstract of the article \cite{liu2018genome} as input to \texttt{o1-preview}, with the following prompt:
\begin{quote}
    ``We are studying a Human foreskin fibroblast cell line from a previously published paper. Here is the abstract of the paper: [insert abstract from \cite{chen2015functional}]. What are the most important biomarker genes for this cell line?"
\end{quote}
From this prompt, which includes additional biological context, \texttt{o1-preview} generates a specific and actionable list of potential biomarkers, similar to the response provided by BRAD (page 12).
However, the output of \texttt{o1-preview} has several shortcomings:  
\begin{itemize}  
    \item \textbf{Data-Centric Limitations:} While \texttt{o1-preview} suggests plausible biomarkers for the cell type studied in \cite{liu2018genome}, it does not account for two critical factors: (1) whether the proposed biomarkers are present in the user's dataset, and (2) whether the dataset contains significant noise or errors in the measurements for the suggested genes.  
    \item \textbf{Lack of Transparency:} Users cannot discern the rationale behind the proposed biomarkers, making it difficult to evaluate the validity or relevance of the suggestions provided by \texttt{o1-preview}.  
    \item \textbf{Reproducibility Concerns:} Due to the absence of transparency in its methodology, the lack of conditioning on the input data, and randomness (temperature) of the model, \texttt{o1-preview} cannot guarantee that it will produce the same list of biomarkers if queried again in the future.  
    \item \textbf{Precision and Relevance:} Unlike \texttt{o1-preview}, which outputs extensive and generic information that a domain expert is likely already familiar with, BRAD provides concise, targeted responses that are more relevant to the user’s needs.  
\end{itemize}

\clearpage
\begin{myboxVS}{\footnotesize{Execute a biomarker selection pipeline for RNAseq data}}
\begin{tcbraster}[raster valign=top, raster halign=center,raster columns=1, raster rows=1,
colframe=white,colback=white,colbacktitle=blue!50!white,
]
\begin{tcolorbox}[title = BRAD's Software Tool with \texttt{gpt-3.5-turbo-0125},colback=blue!5!white,colframe=blue!75!black,fonttitle=\bfseries, width = 6cm]
\footnotesize
\begin{flushright}
\textbf{Software Tool:} Execute data loading script
\end{flushright}

\textbf{LLM:} Based on the software output, the 2015 gene expression dataset with gene coordinates has been successfully converted and saved to the following location: /home/jpic/BRAD-Examples/DMD-Biomarkers/output/November 27, 2024 at 01:14:17 PM/S0-2015.pkl. This dataset is shaped as (19393, 9, 2) where the modes represent genes by time points by experimental replicates. You can load this dataset from the specified file path to access the gene expression data with gene coordinates from 2015.

\vspace{2mm}

\begin{flushright}
\textbf{Software Tool:} Execute modeling script
\end{flushright}

\textbf{LLM:} The dynamics model of the 2015 gene coordinate time series data has been built using Dynamic Mode Decomposition (DMD). The output file for this model is saved at /home/jpic/BRAD-Examples/DMD-Biomarkers/output/November 27, 2024 at 01:14:17 PM/S1-2015.pkl. The input file used for this model was /home/jpic/BRAD-Examples/DMD-Biomarkers/output/November 27, 2024 at 01:14:17 PM/S0-2015.pkl. The DMD Rank for this model is None, indicating the complexity of the model. You can access and analyze this dynamics model from the specified file path to further explore the gene coordinate time series data from 2015.

\vspace{2mm}

\begin{flushright}
\textbf{Software Tool:} Execute observability script
\end{flushright}

\textbf{LLM:} The biomarker selection process has been performed on the model of the 2015 dataset using Dynamic Mode Decomposition (DMD) and Observability Energy guided sensor selection. The sensor file containing the selected biomarkers has been saved at /home/jpic/BRAD-Examples/DMD-Biomarkers/output/November 27, 2024 at 01:14:17 PM/S3-2015.csv. 

The top 30 ranked genes based on their contribution to observability are listed below:
\begin{enumerate}
    \item MT-CO1
    \item ACTB
    \item MT-ND4
    \item MT-CO3
    \item MT-CO2
    \item MT-ND5
    \item ACTG1
    \item LCP1
    \item B2M
    \item MT-CYB
    \item CD74
    \item GAPDH
    \item TUBB
    \item MT-ATP6
    \item ENO1
    \item HLA-DRA
    \item EEF2
    \item PKM
    \item HSPA8
    \item RPS6
    \item MT-ND4L
    \item HNRNPA2B1
    \item TPT1
    \item LDHA
    \item MT-ND1
    \item HSP90AB1
    \item MT-ND6
    \item HSP90AA1
    \item TMSB4X
    \item SRRM2
\end{enumerate}
These genes have been ranked based on their importance as biomarkers in the dataset. You can further analyze and interpret the results by referring to the specified file path for the sensor selection output.
\end{tcolorbox}
\end{tcbraster}
\end{myboxVS}

\clearpage
\begin{myboxVS}{\footnotesize{Execute a biomarker selection pipeline for RNAseq data}}
\begin{tcbraster}[raster valign=top, raster halign=center,raster columns=1, raster rows=1,
colframe=white,colback=white,colbacktitle=blue!50!white,
]
\begin{tcolorbox}[title = OpenAI chatbot with \texttt{o1-preview},colback=blue!5!white,colframe=blue!75!black,fonttitle=\bfseries, width = 6cm]
\footnotesize
Selecting biomarkers from RNAseq data involves a multi-step pipeline designed to identify genes or transcripts that can serve as reliable indicators for specific conditions or biological states. Below is a typical framework to execute such a pipeline:

\textbf{Step 1: Load and Explore Data}
\begin{enumerate}
    \item Input Data:
    \begin{itemize}
        \item RNAseq data matrix (e.g., gene expression counts).
        \item Associated metadata (e.g., phenotypes, conditions, or clinical outcomes).
    \end{itemize}
    \item Preprocessing:
    \begin{itemize}
        \item Normalize raw counts (e.g., TPM, RPKM, or using a tool like DESeq2 for variance stabilization).
        \item Perform exploratory data analysis (EDA) to check for batch effects, outliers, and data distribution.
    \end{itemize}
\end{enumerate}

\textbf{Step 2: Preprocess and Normalize Data}
\begin{itemize}
    \item Quality Control:
\begin{itemize}
    \item Filter low-count genes (e.g., genes expressed in <5\% of samples or with counts below a threshold).
    \item Remove unwanted variation (batch effect removal using tools like sva or Combat).
\end{itemize}
\item Normalization:
\begin{itemize}
    \item Use appropriate normalization methods such as:
    \begin{itemize}
        \item `TMM` (Trimmed Mean of M-values) for edgeR.
        \item `VST` (Variance Stabilizing Transformation) for DESeq2.
    \end{itemize}
\end{itemize}
\end{itemize}

\textbf{Step 3: Identify Differentially Expressed Genes (DEGs)}
\begin{itemize}
    \item Perform statistical tests to find DEGs between conditions or phenotypes:
    \begin{itemize}
        \item DESeq2 or edgeR for count data.
        \item Adjust for multiple testing using methods like FDR (False Discovery Rate) to identify significant genes (e.g., FDR < 0.05).
    \end{itemize}
\end{itemize}

\textbf{Step 4: Dimensionality Reduction}
\begin{itemize}
    \item Use techniques like PCA (Principal Component Analysis) or t-SNE to reduce dimensionality and visualize the most variable genes across conditions.
\end{itemize}

\textbf{Step 5: Feature Selection for Biomarker Discovery}
\begin{itemize}
    \item Apply feature selection techniques to pinpoint biomarkers:
    \begin{itemize}
        \item Filter Methods:
        \begin{itemize}
            \item Select top N genes based on fold change or p-value.
            \item Use variance thresholds to filter less variable genes.
        \end{itemize}
        \item Wrapper Methods:
        \begin{itemize}
            \item Recursive Feature Elimination (RFE) with classifiers like SVM or Random Forest.
        \end{itemize}
        \item Embedded Methods:
        \begin{itemize}
            \item LASSO (Least Absolute Shrinkage and Selection Operator).
            \item Elastic Net Regression.
        \end{itemize}
    \end{itemize}
\end{itemize}

\textbf{Step 6: Model Building and Validation}
\begin{enumerate}
    \item Model Selection:
    \begin{itemize}
        \item Train machine learning models (e.g., Random Forest, Support Vector Machines) on the selected features.
        \item Use cross-validation to assess performance (e.g., accuracy, AUC-ROC).
    \end{itemize}
    \item Biological Pathway Enrichment:
    \begin{itemize}
        \item Analyze the selected biomarkers for enrichment in known pathways using tools like GSEA, Reactome, or KEGG.
    \end{itemize}
\end{enumerate}

\textbf{Step 7: Validation of Biomarkers}
\begin{itemize}
    \item Validate the identified biomarkers on an independent dataset (if available) to ensure robustness.
    \item Optionally, validate the findings experimentally (e.g., qPCR or other assays).
\end{itemize}

\textbf{Implementation in Code}
Would you like to focus on a specific programming environment (e.g., Python, R)? If you already have a dataset, I can implement a specific pipeline in the chosen language.
\end{tcolorbox}
\end{tcbraster}
\end{myboxVS}

\begin{myboxVS}{\parbox{12cm}{\footnotesize We are studying a Human foreskin fibroblast cell line from a previously published paper. Here is the abstract of the paper: [insert abstract from \cite{chen2015functional}]. What are the most important biomarker genes for this cell line?}}
\begin{tcbraster}[raster valign=top, raster halign=center,raster columns=1, raster rows=1,
colframe=white,colback=white,colbacktitle=blue!50!white]
\begin{tcolorbox}[title = OpenAI chatbot with \texttt{o1-preview},colback=blue!5!white,colframe=blue!75!black,fonttitle=\bfseries, width = 6cm]
\footnotesize
The abstract you provided describes a study focused on transcription factor (TF)-mediated reprogramming of human fibroblasts, particularly looking at the genome architecture and how it reorganizes during the process of transitioning these cells into a myogenic (muscle-related) identity. The key finding is that genome architecture reorganization precedes transcriptional activation of the myogenic program. This suggests that the study investigates how changes in the 3D structure of chromatin influence gene expression during cellular reprogramming.
\vspace{2mm}

Based on the study's focus, the most important biomarker genes for this human foreskin fibroblast cell line would be associated with:

\begin{itemize}
    \item \textbf{Fibroblast Identity:} Baseline markers to confirm the fibroblast nature of the starting cells.
    \item \textbf{Myogenic Reprogramming:} Markers that identify the activation of myogenic transcriptional programs.
    \item \textbf{Genome Architecture and Transcription Factors:} Genes involved in chromatin remodeling and transcriptional regulation, specifically those that play a role in reprogramming.
\end{itemize}

Here are the key categories and corresponding genes likely to be important as biomarkers for this cell line:

\vspace{1mm}
\textbf{1. Fibroblast Identity Biomarkers}
\vspace{1mm}

\noindent
These genes are important to confirm the starting identity of the human foreskin fibroblasts before reprogramming begins:

\begin{itemize}
    \item \textbf{VIM (Vimentin):} A marker for fibroblast identity. It is expressed in mesenchymal cells and is a fundamental marker to confirm the fibroblast origin.
    \item \textbf{S100A4 (Fibroblast-Specific Protein 1, FSP1):} Specific to fibroblasts and used to ensure that the starting cell line retains fibroblast characteristics.
    \item \textbf{COL1A1 / COL1A2 (Collagen Type I Alpha 1 and 2):} These genes encode the major components of the extracellular matrix in fibroblasts, essential for confirming fibroblast functionality.
    \item \textbf{FN1 (Fibronectin 1):} This extracellular matrix protein gene is a hallmark of fibroblast function, particularly in wound healing and tissue remodeling.
\end{itemize}

\vspace{1mm}
\textbf{2. Myogenic Program Biomarkers}
\vspace{1mm}

\noindent
The study focuses on myogenic reprogramming, so genes involved in myogenic lineage development are critical:

\begin{itemize}
    \item \textbf{MYOD1 (Myogenic Differentiation 1):} This master regulator is a key transcription factor involved in initiating myogenic differentiation. It plays a critical role in reprogramming fibroblasts into muscle cells.
    \item \textbf{MYOG (Myogenin):} Another essential transcription factor, MYOG, is involved in the terminal differentiation of myoblasts into muscle cells and is crucial for the commitment to a myogenic fate.
    \item \textbf{MYF5 (Myogenic Factor 5):} A gene involved in the early stages of myogenic lineage commitment, working alongside MYOD1 to initiate muscle-specific gene expression.
    \item ...
\end{itemize}

\vspace{1mm}
\textbf{3. Transcriptional Regulators and Genome Architecture Biomarkers} ...
\vspace{1mm}

\vspace{1mm}
\textbf{4. Chromatin Dynamics and Genome Topology Markers} ...
\vspace{1mm}

\vspace{1mm}
\textbf{5. Rhythmic and Biological Rhythms-Related Genes} ...
\vspace{1mm}

\vspace{1mm}
\textbf{Summary of Key Biomarker Genes for This Cell Line}
\vspace{1mm}
\begin{itemize}
    \item \textbf{Fibroblast Identity:} VIM, S100A4, COL1A1, FN1
    \item \textbf{Myogenic Reprogramming:} MYOD1, MYOG, MYF5, PAX7, DES
    \item \textbf{Chromatin and Transcriptional Regulation:} CTCF, SMARCA4, EP300, EZH2, GATA4/6
    \item \textbf{Genome Architecture and Chromatin Dynamics:} RAD21, SCC1, TP53
    \item \textbf{Rhythmic Genes:} CLOCK, ARNTL
\end{itemize}

These genes collectively serve as crucial biomarkers for tracking fibroblast identity, myogenic reprogramming, chromatin remodeling, and the interplay between genome architecture and transcriptional programs in this cell line.
\end{tcolorbox}
\end{tcbraster}
\end{myboxVS}

\end{document}